\documentclass[preprint,review,10pt]{elsarticle}
\usepackage[english]{babel}
\usepackage[T1]{fontenc}
\usepackage{times}
\usepackage{etoolbox}
\makeatletter

\patchcmd{\ps@pprintTitle}{\footnotesize\itshape
      Preprint submitted to \ifx\@journal\@empty Elsevier
      \else\@journal\fi\hfill\today}{\scriptsize{Preprint submitted to ASOC \hfill \today}}{}{}

\usepackage{multicol}
\usepackage{float}
\usepackage{natbib} 
\usepackage{graphicx}
\usepackage[lined,  boxed,  linesnumbered,  ruled]{algorithm2e}
\graphicspath{{Figures/}}
\usepackage{lineno,hyperref}
\usepackage{amssymb, amsfonts, amsmath, graphicx, epstopdf, bbding, lipsum, multirow, makecell, picinpar, enumerate}
\usepackage[table]{xcolor}
\usepackage{colortbl}
\usepackage{xspace}
\usepackage{subfigure} 
\usepackage{xcolor}
\usepackage[normalem]{ulem} 

\makeatother
\usepackage{comment}
\usepackage{hyperref}
\usepackage{amsmath}

\usepackage{amsmath,amssymb,amsfonts}
\usepackage{algorithmic}
\usepackage[]{graphicx}
\graphicspath{{Figures/}}
\usepackage{lipsum}
\usepackage{balance}
\usepackage{textcomp}
\usepackage{xcolor}
\usepackage{multirow}
\usepackage{color}
\usepackage{lscape}
\usepackage{longtable,booktabs}
\usepackage[numbers]{natbib} 

\begin{document}

\begin{frontmatter}

\title{TrafficKAN-GCN: Graph Convolutional-based Kolmogorov-Arnold Network for\\ Traffic Flow Optimization}

\author[add1]{Jiayi~Zhang\corref{contrib}}
\ead{smyjz19@nottingham.edu.cn}
\author[add2]{Yiming~Zhang\corref{contrib}}
\ead{scyyz41@nottingham.edu.cn}
\author[add1]{Yuan~Zheng}
\ead{smyyz25@nottingham.edu.cn}
\author[add4]{Yuchen~Wang}
\ead{yuchwang@uw.edu}
\author[add8]{Jinjiang~You}
\ead{jinjiany@alumni.cmu.edu} 
\author[add1]{Yuchen Xu}
\ead{smyyx14@nottingham.edu.cn}
\author[add3]{Wenxing~Jiang}
\ead{ssywj4@nottingham.edu.cn}

\author[add5,add6]{Soumyabrata~Dev\corref{mycorrespondingauthor}}

\address[add1]{School of Mathematics and Applied Mathematics, University of Nottingham Ningbo China}
\address[add2]{School of Computer Science, University of Nottingham Ningbo China}
\address[add4]{Department of Electrical and Computer Engineering, University of Washington}
\address[add8]{School of Computer Science, Carnegie Mellon University}
\address[add3]{School of Engineering, University of Nottingham Ningbo China}
\address[add5]{School of Computer Science, University College Dublin}
\address[add6]{The ADAPT SFI Research Centre}
\cortext[contrib]{Authors contributed equally to this research.}
\cortext[mycorrespondingauthor]{Corresponding author. Tel.: + 353 1896 1797.}
\ead{soumyabrata.dev@ucd.ie}

\begin{abstract}
Urban traffic optimization is crucial for improving transportation efficiency and alleviating congestion. Traditional algorithms like Dijkstra’s and Floyd’s, along with Genetic Algorithms (GA), are widely used for route planning but struggle with large-scale, dynamic traffic networks due to computational complexity and limited ability to capture nonlinear spatial-temporal dependencies. To address these challenges, we propose a hybrid framework combining Kolmogorov-Arnold Networks (KAN) with Graph Convolutional Networks (GCN) for urban traffic flow optimization. KAN’s ability to approximate nonlinear functions, along with GCN’s capacity to model spatial dependencies in graph-structured data, provides a more effective representation of traffic dynamics. We evaluate KAN-GCN on real-world datasets and compare it to traditional models like Multi-Layer Perceptrons (MLP) and decision trees. While MLP-GCN shows higher accuracy, KAN-GCN excels in handling noisy and complex traffic patterns. Results highlight challenges in applying KAN to large-scale networks and suggest potential improvements. Additionally, we explore extending the framework to Transformer models for scalable and dynamic traffic forecasting. Our findings indicate that combining KAN with GCN offers promising potential for intelligent traffic management, with further optimization possible for large-scale applications. 
\end{abstract}

\begin{keyword}
Traffic Network Optimization, Graph Convolutional Networks, Kolmogorov–Arnold Networks, Machine Learning, Predictive Analytics, Network Forecasting
\end{keyword}
\end{frontmatter}

\section{Introduction}
\begin{figure}[h]
    \centering
    \includegraphics[width=0.9\linewidth]{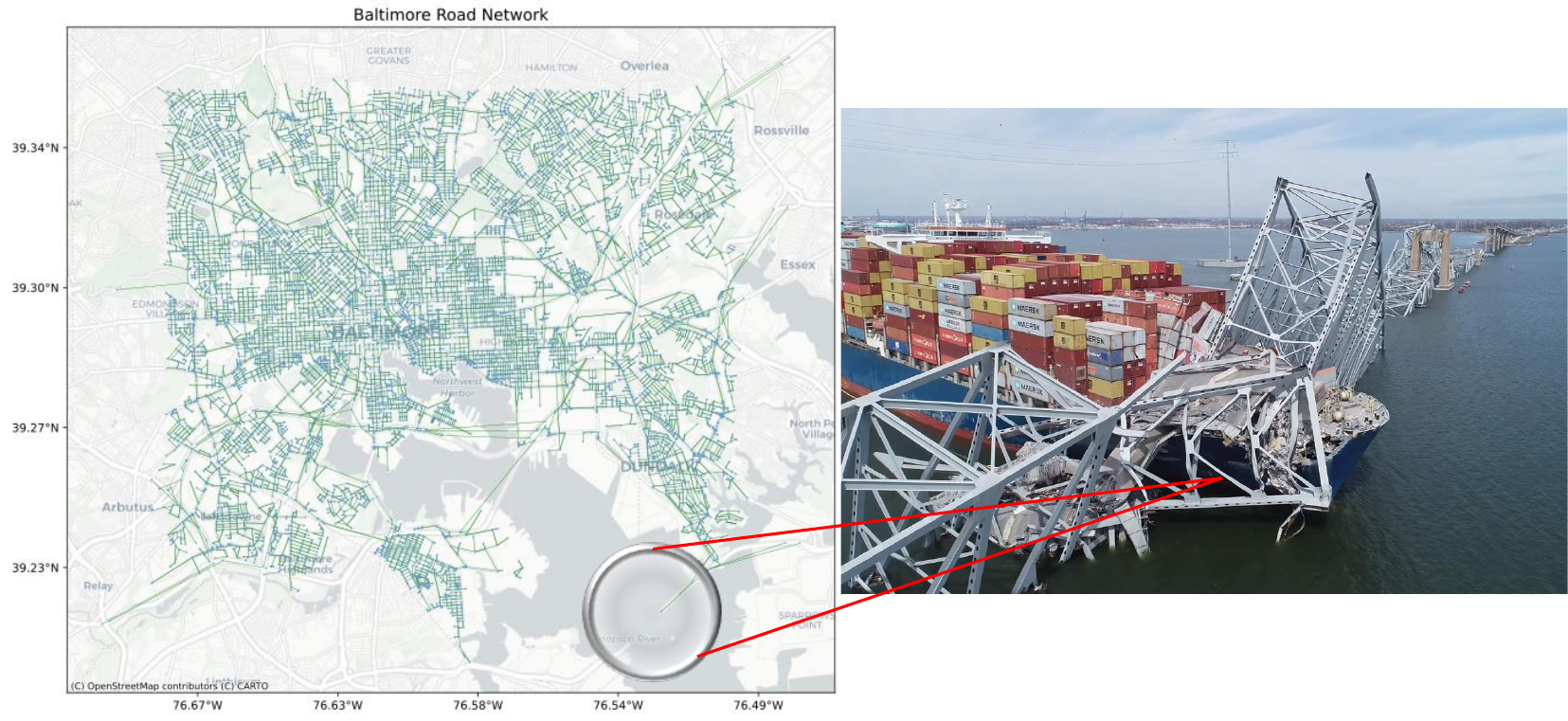} 
    \caption{Baltimore's transportation network, showing major roads and highways (Left). The Francis Scott Key Bridge collapse on March 26, 2024, which disrupted traffic and economic activities (Right).}
    \label{fig:key_bridge_collapse}
\end{figure}

On March 26, 2024, the Francis Scott Key Bridge in Baltimore collapsed after being struck by the container ship \textit{Dali} (Figure~\ref{fig:key_bridge_collapse}). This catastrophic event led to the tragic deaths of six construction workers and caused severe disruptions to both local traffic and the broader supply chain~\cite{National}, as the bridge served as a critical artery for transportation and commerce in the region. The collapse led to significant congestion on alternative routes, including the I-95 and I-895 highways, exacerbating traffic bottlenecks and increasing travel times for both commuters and freight transport~\cite{AFRO}. The collapse of such a vital infrastructure component underscores the pressing need for effective traffic network optimization strategies, particularly in urban environments where transportation resilience and efficiency are paramount. In response to such disruptions, robust and adaptive traffic management solutions are required to mitigate congestion and ensure the smooth operation of urban transportation systems~\cite{li2021t}.

Traditional graph-based mathematical models~\cite{holland1992adaptation}, such as Dijkstra's algorithm~\cite{dijkstra1959note}, Floyd's algorithm~\cite{floyd1962algorithm}, and Genetic Algorithms (GA)~\cite{5701841}, have long been utilized for route planning and traffic optimization~\cite{SUMO2018}. While these models are effective in static environments~\cite{pallottino1984shortest}, they often struggle with dynamic, large-scale traffic systems due to computational constraints and their inability to adapt to real-time fluctuations in road conditions~\cite{nannicini2012bidirectional}. As urban road networks continue to grow in complexity, there is a pressing need for data-driven models capable of learning and predicting traffic patterns dynamically~\cite{wang2021usage}. Graph learning has a wide range of applications, from recommendation systems~\cite{xu2024aligngroup,xu2024mentor} and music influence analysis~\cite{WANG2024100601} to path planning~\cite{diao2023graphneuralnetworkbased,liu2025graphneuralnetworkstravel}, and our work is closely related to path planning and urban traffic optimization. Recent advancements in deep learning and graph-based machine learning have introduced novel methodologies to address these challenges~\cite{bronstein2017geometric}. In particular, Graph Convolutional Networks (GCNs) have shown strong potential in modeling traffic networks, leveraging their ability to learn spatial dependencies from graph-structured data~\cite{Zhao2020TGCN}. However, conventional GCN-based approaches still face challenges in capturing nonlinear relationships and handling the highly dynamic nature of traffic conditions~\cite{zhang2019graphcn}.

To ameliorate the aforementioned issues, we propose a hybrid optimization framework that integrates KAN with GCN to optimize urban traffic flow~\cite{kiamari2024gkangraphkolmogorovarnoldnetworks,dong2024kolmogorovarnoldnetworkskantime,decarlo2024kolmogorovarnoldgraphneuralnetworks}. KAN~\cite{decarlo2024kolmogorovarnoldgraphneuralnetworks,Padmavathi2024MarineTraffic}, a recent advancement in neural network architectures, offers superior expressiveness and function approximation capabilities, making it well-suited for capturing complex, nonlinear traffic dynamics. By combining the structural learning capabilities of GCN with the adaptive approximation power of KAN, our proposed model aims to enhance the robustness and flexibility of traffic optimization strategies. The main contributions of our method are threefold:

\begin{itemize}
    \item In our comparison with conventional models, we conducted a comprehensive evaluation of the KAN-GCN model against traditional machine learning approaches, such as MLP~\cite{kong2022exploring} and decision trees, using real-world traffic datasets. While MLP-GCN demonstrated superior prediction accuracy, KAN-GCN showed significant advantages in its robustness to noisy data and its ability to adapt to complex traffic patterns~\cite{dong2024kolmogorovarnoldnetworkskantime, Bresson2024KAGNNs}.
    
    \item Our study provides valuable insights into the challenges and opportunities of integrating KAN into urban traffic networks, highlighting potential avenues for future research to enhance traffic optimization models and strategies.
    
    \item We propose extending the KAN-GCN framework by incorporating Transformer-based architectures for time-series traffic forecasting. This extension underscores the scalability of the model and its ability to address the dynamic and large-scale challenges inherent in traffic forecasting.
\end{itemize}

This paper is organized as follows: Section 2 presents a comprehensive review of related work, starting with an analysis of graph-based traffic network models and proceeding to discuss the integration of KAN with these models. Section 3 details the dataset and the assumptions made for the study. In Section 4, we outline the data preprocessing process, including feature engineering and selection techniques. Section 5 introduces the methodology, covering traffic network modeling, the KAN-GCN architecture, loss function, and training strategy, as well as stakeholder preferences and their incorporation into the model. Section 6 provides an impact analysis on different stakeholders, and Section 7 presents the experimental evaluation, including the evaluation metrics and quantitative experiments. Finally, Section 8 concludes the paper and offers insights for future research directions.

\section{Related Work}

\subsection{Graph-Based Traffic Network Optimization}

Traditional traffic network optimization has heavily relied on graph-based mathematical models, such as Dijkstra's Algorithm and the Floyd-Warshall Algorithm, which, although effective in static scenarios, face significant challenges in dynamic urban environments. Dijkstra's Algorithm, with \(O(n \log n)\) complexity, is efficient for small-scale networks but becomes computationally expensive in large-scale applications~\cite{dijkstra1959note}. The Floyd-Warshall Algorithm, which computes shortest paths for all node pairs, suffers from \(O(n^3)\) complexity, rendering it impractical for real-time traffic optimization in large-scale networks~\cite{floyd1962algorithm}. Genetic Algorithms (GA) have also been used to optimize traffic flow by simulating evolutionary selection processes~\cite{holland1992adaptation}, but their high computational cost and slow convergence make them less suitable for real-time applications. Recent studies have explored the use of graph-based models in other fields, such as Xu et al.'s work on AlignGroup, which leverages graph-based techniques for group recommendation~\cite{xu2024aligngroup,xu2024mentor}, and the MFCSNet model, which applies graph-based social network analysis to measure musical influence~\cite{WANG2024100601}. These examples demonstrate the potential of graph-based models, when combined with machine learning, to optimize traffic systems by aligning stakeholder preferences and leveraging dynamic data, offering a promising solution to overcome the limitations of traditional optimization methods in traffic flow management.

With the rise of deep learning, data-driven approaches have significantly enhanced traffic network modeling. Recurrent Neural Networks (RNNs), Long Short-Term Memory (LSTM) networks, and Gated Recurrent Units (GRUs) have been widely employed for short-term traffic flow prediction, capturing sequential dependencies in traffic data~\cite{Polson2017TrafficDL}. However, these models face challenges in modeling long-range dependencies and suffer from vanishing gradient issues, which hinder their performance in large-scale networks~\cite{pascanu2013trainingrnn}. To overcome these limitations, Graph Neural Networks (GNNs)~\cite{wu2021gnnsurvey,Cai2020TrafficTransformer} have gained traction as a powerful tool for traffic modeling, thanks to their ability to capture spatial dependencies inherent in road networks. For instance, Yu \textit{et al.} (2018) introduced Spatio-Temporal Graph Convolutional Networks (ST-GCN), which combine graph convolutions with temporal dependencies to improve traffic forecasting~\cite{yu2018spatio}. Additionally, Li \textit{et al.} (2018) proposed the Diffusion Convolutional Recurrent Neural Network (DCRNN), which integrates diffusion graph convolutions to model spatial traffic dynamics while employing recurrent structures to capture temporal trends~\cite{li2018diffusion}. Despite these advancements, GCN-based methods still face challenges, such as computational overhead from adjacency matrix normalization, limitations in nonlinear function approximation, and the need for additional architectures like RNNs or Transformers to effectively model temporal dependencies~\cite{li2018deeper}. Additionally, Convolutional Neural Networks (CNNs), traditionally used for image processing tasks, have also been explored for traffic prediction, leveraging their ability to capture spatial patterns in traffic data, similar to how they process visual features in images~\cite{ma2017learning}.

Recent developments in traffic prediction have highlighted the effectiveness of combining graph convolutional networks (GCNs) with other machine learning models for network optimization. For example, T-GCN integrates temporal and spatial features for improved forecasting~\cite{li2025cp2m}, while DDUNet introduces segmentation capabilities adaptable to traffic patterns~\cite{li2025ddunetdualdynamicunet,10640450}. Majidiparast \textit{et al.}~\cite{railway} proposed a GCN-based method for predictive maintenance in railway systems, but such models often remain limited to local dependencies. Similarly, Akram \textit{et al.}~\cite{akram2024fuzzy} demonstrated how integrating preference modeling with multi-criteria decision analysis (MCDA) enhances decision robustness under uncertainty. Building on these insights, our KAN-GCN model incorporates stakeholder-specific preferences while addressing the limitations of existing GCNs through global function approximation, offering a more flexible and scalable framework for traffic network optimization under dynamic and uncertain conditions.

Recently, attention-based models have been introduced to enhance spatial information aggregation in graph structures. The Graph Attention Network (GAT) improves upon GCNs by assigning different importance weights to neighboring nodes using a self-attention mechanism, making it more effective in heterogeneous traffic networks where node connectivity varies significantly~\cite{velivckovic2017graph}. However, the computational cost of GAT increases due to its attention mechanism, limiting its scalability for large-scale real-time traffic modeling. Transformer-based architectures have also been explored for traffic forecasting, offering superior performance in long-horizon predictions by leveraging self-attention to capture complex temporal dependencies. While these models exhibit strong predictive accuracy, their high computational complexity remains a barrier to real-time deployment.

\subsection{KAN and Their Integration with GCNs}

KAN has recently emerged as a novel deep learning paradigm, inspired by the Kolmogorov-Arnold representation theorem, which states that any continuous multivariable function can be decomposed into a finite set of univariate functions~\cite{kurkova1992kolmogorov}. Unlike traditional MLPs that rely on fixed activation functions, KAN employs learnable univariate transformations, enabling more expressive function approximations. This adaptive activation mechanism allows KAN to model complex nonlinear relationships more effectively, making it particularly suitable for dynamic systems like urban traffic networks~\cite{KAGNNs2024}.

Integrating KAN with GCNs presents an opportunity to enhance the nonlinear modeling capability of graph-based architectures. Standard GCNs primarily rely on linear feature propagation, limiting their ability to capture highly nonlinear traffic dynamics. By incorporating KAN into GCN architectures, the model gains the flexibility to learn adaptive transformation functions, potentially improving feature expressiveness and model interpretability. However, the application of KAN in graph-based traffic modeling remains largely unexplored. One of the primary challenges of KAN-GCN integration is its computational cost, as the learnable activation functions introduce additional complexity. Moreover, KAN-based models require longer training epochs compared to standard GCNs, which may affect real-time performance.

This study proposes a hybrid optimization framework that integrates KAN with GCN to improve urban traffic modeling. The key contributions include the integration of KAN with GCN architectures to enhance nonlinear representation and spatial-temporal adaptability, providing a more flexible and interpretable traffic modeling framework. Additionally, we conduct a comparative analysis of KAN-GCN against traditional methods such as MLP-GCN and GA, evaluating prediction accuracy, computational efficiency, and robustness to real-world traffic variations. Finally, we explore the potential extension of KAN-GCN with Transformer-based architectures to improve long-term traffic forecasting, demonstrating its scalability and applicability in complex urban transportation networks.

\section{Dataset \& Assumptions}

The dataset used in this study is sourced from the Maryland Department of Transportation (MDOT), the Baltimore Metropolitan Council’s Regional GIS Data Center, and the Baltimore County Open Data portal. \footnote{\url{https://baltometro.org/about-us/datamaps/regional-gis-data-center}} \footnote{\url{https://opendata.baltimorecountymd.gov}} It includes:

\begin{itemize}
    \item \textbf{Road Network Data}: Information on road segments, intersections, and connectivity.
    \item \textbf{Traffic Volume Data}: Annual average daily traffic (AADT) counts for key road segments.
    \item \textbf{Public Transit Data}: Bus routes, bus stops, and transit schedules.
    \item \textbf{Spatial and GIS Data}: Geographic coordinates and mapping layers.
\end{itemize}

The study makes the following assumptions: the road network is modeled as an undirected graph, with nodes representing intersections and edges representing road segments, where traffic flow is assumed to be uniform and roads have limited capacity. It is assumed that drivers always select the optimal path based on the available network, and the impact of extreme weather conditions on traffic is not considered. Additionally, the dataset is regarded as reliable, with no significant errors or missing data. This dataset provides crucial temporal and spatial attributes for traffic flow prediction and urban mobility optimization.

\section{Data Preprocessing and Feature Engineering}

\subsection{Data Preprocessing}

To ensure the reliability and consistency of the traffic data, a systematic preprocessing pipeline was applied. The dataset, sourced from multiple government and transportation agencies, including the Baltimore City Open Data Portal, Maryland Department of Transportation (MDOT SHA~\cite{Leduc2008TrafficData}), and Baltimore Metropolitan Council (BMC), contains traffic volume records (AADT), road network topology, and public transit data. Given the presence of missing values, anomalies, and noise in real-world traffic data, we adopted a structured approach to data cleaning.

For missing values, we applied different strategies based on the proportion of missing entries. If missing values constituted less than \(5\%\), K-Nearest Neighbors (KNN)~\cite{inproceedings} imputation was used to estimate missing values. When missing data exceeded \(30\%\), the affected samples were removed to prevent bias in model learning. Additionally, anomaly detection was conducted using Z-score~\cite{patro2015normalizationpreprocessingstage} normalization, where data points exceeding \( |Z| > 3 \) were classified as outliers and replaced using a rolling mean smoothing method:

\begin{equation}
Z = \frac{x - \mu}{\sigma}
\end{equation}
where \( \mu \) and \( \sigma \) represent the mean and standard deviation, respectively.

Considering the varying scales of different traffic-related attributes, the min-max normalization method is used to ensure that all features are within the normalized range of  \( [0,1] \), thus improving the convergence of the model:
\begin{equation}
X' = \frac{X - X_{\min}}{X_{\max} - X_{\min}}
\end{equation}

\subsection{Feature Engineering}

To enhance the performance of the KAN-GCN in predicting traffic flow, we extracted spatial and temporal features that capture network topology and time-dependent patterns.

\textbf{Traffic Network Representation.} The transportation network is formulated as an undirected weighted graph \( G = (V, E) \), where \( V \) corresponds to intersections and bus stops, while \( E \) denotes road segments connecting them. Each edge \( e_{ij} \) is assigned a weight:

\begin{equation}
w_{ij} = \alpha L_{ij} + \beta S_{ij} + \gamma C_{ij} + \delta T_{ij}
\label{eqn.weight}
\end{equation}
where \( L_{ij} \) represents road length, \( S_{ij} \) is the speed limit, \( C_{ij} \) denotes congestion level derived from real-time traffic data, and \( T_{ij} \) corresponds to estimated travel time based on historical records. The parameters \( \alpha, \beta, \gamma, \delta \) are trainable coefficients controlling the contribution of each factor. To maintain numerical stability, the adjacency matrix is normalized as follows:

\begin{equation}
\tilde{A} = D^{-\frac{1}{2}} A D^{-\frac{1}{2}}
\label{111}
\end{equation}
where \( D \) is the diagonal degree matrix of the graph to represent the number of connections (edges) per node, whose normalization helps prevent exploding gradients and ensures efficient learning in the GCN:

\begin{equation}
D_{ii} = \sum_j A_{ij}
\label{222}
\end{equation}

\textbf{Temporal Features~\cite{Taylor2010ConvSpatioTemporal}.} To capture temporal dependencies in traffic flow, we incorporate historical traffic observations and periodic time indicators. A moving average function aggregates past \( t \) time steps:

\begin{equation}
X_t^{(avg)} = \frac{1}{t} \sum_{k=1}^{t} X_{t-k}
\end{equation}

Moreover, cyclical time encoding is introduced using sine and cosine functions to preserve temporal periodicity:

\begin{equation}
T_{sin} = \sin \left(\frac{2\pi T}{24}\right), \quad T_{cos} = \cos \left(\frac{2\pi T}{24}\right)
\end{equation}
where \( T \) represents the hour of the day.

\textbf{Road Attributes~\cite{Jan2018CNN_RoadAttributes}.} Each road segment is characterized by lane count, road type (highway, arterial, or secondary road), and bus stop density. Bus stop density is computed as:

\begin{equation}
D_{bus} = \frac{N_{bus}}{L_{road}}
\end{equation}
where \( N_{bus} \) denotes the number of bus stops along a given road segment, and \( L_{road} \) represents the total length of the road.

\subsection{Feature Selection}

To enhance model efficiency while retaining critical predictive information, we applied feature selection methods based on Mutual Information (MI) and SHapley Additive exPlanations (SHAP).

\textbf{Mutual Information Selection~\cite{IAM}.} MI quantifies the dependency between each feature and the target variable (traffic flow):

\begin{equation}
MI(X, y) = \sum_{x,y} p(x, y) \log \frac{p(x, y)}{p(x)p(y)}
\end{equation}
where higher MI values indicate stronger predictive relevance.

\textbf{SHAP-Based Feature Importance~\cite{SHAP}.} SHAP values were computed to evaluate the impact of individual features on model predictions:

\begin{equation}
\Phi_i = \mathbb{E}[f(X) | X_i] - \mathbb{E}[f(X)]
\end{equation}
where \( \Phi_i \) represents the contribution of feature \( X_i \). Based on SHAP and MI evaluations, the final selected features include historical traffic data, road topology, time-of-day indicators, road attributes (lane count, speed limit), and bus stop density.

\section{Methodology}

\subsection{Traffic Network Modeling}
To model urban traffic dynamics, we represent the transportation system as a weighted graph \( G = (V, E, W) \), where \( V \) is the set of nodes representing key intersections or transportation hubs, and \( E \) is the set of edges representing road segments between nodes. The graph's edge weights \( w_{ij} \) incorporate relevant traffic attributes, including road length \( L_{ij} \), speed limit \( S_{ij} \), congestion level \( C_{ij} \), and estimated travel time \( T_{ij} \), which can be expressed as Eqn.~\eqref{eqn.weight}. To ensure stable learning and effective message passing in the GCN~\cite{Wu2021GCN-LSTM}, the adjacency matrix \( A \) of the graph is normalized as Eqn.~\eqref{111} and Eqn.~\ref{222}, capturing spatial dependencies between different road segments and allowing GCN to effectively model the traffic network structure. Unlike Convolutional Neural Networks (CNNs), which are typically used to process grid-like data such as images~\cite{BATRA2022200039, WANG2022102243, LI2025101536, WANG2021}, GCNs are designed to handle graph-structured data, making them more suitable for representing complex relationships in traffic networks~\cite{chen2019gated}. CNN-based dense feature extractors have also been widely used in various domains and applications, including medical prediction~\cite{DEV2022100032, tang2024optimized, pan2024accurate}, perception and calibration~\cite{WANGCAR2022, you2025multi, zhenqi23}, remote sensing~\cite{cui2024superpixel, li2025segregation}, saliency object detection~\cite{li2023daanet, Li_2024_BMVC}, as well as robotics \cite{zhu2023fanuc, Lin2024JointPT, Huo2023HumanorientedRL}. However, GCNs provide the advantage of modeling non-Euclidean data, where relationships between data points are not spatially structured in grids, enabling better handling of the spatial dependencies in urban traffic systems. In the context of KAN-GCN, the graph convolution operations are further enhanced by integrating KAN, which introduces adaptive non-linear transformations, enabling the model to approximate complex traffic patterns more accurately. To account for dynamic uncertainties in traffic flow, we introduce a transportation-based kernel metric for adaptive risk evaluation. This approach is inspired by recent advances in Markov system modeling using integrated transportation distances~\cite{zhen1ilin23}.

\begin{figure}[htbp]
    \centering
    \begin{minipage}{0.48\textwidth}
        \centering
        \resizebox{\textwidth}{!}{
        \begin{tabular}{|c|c|c|c|c|c|c|}
        \hline
        \rowcolor[HTML]{D3D3D3} 
        \textbf{Start Node} & \textbf{End Node} & \textbf{Road Length (km)} & \textbf{Speed Limit (km/h)} & \textbf{Congestion Level (0-1)} & \textbf{Estimated Travel Time (min)} & \textbf{Weight \( w_{ij} \)} \\
        \hline
        V1 & V2 & 6 & 60 & 0.2 & 6 & 6.3 \\
        \hline
        V1 & V3 & 4 & 50 & 0.3 & 5 & 5.1 \\
        \hline
        V1 & V4 & 8 & 40 & 0.4 & 10 & 10.2 \\
        \hline
        V1 & V5 & 10 & 70 & 0.1 & 7 & 7.1 \\
        \hline
        V2 & V3 & 3 & 60 & 0.3 & 4 & 4.2 \\
        \hline
        V2 & V4 & 6 & 50 & 0.2 & 6 & 6.4 \\
        \hline
        V2 & V5 & 5 & 40 & 0.4 & 5 & 5.3 \\
        \hline
        V3 & V4 & 7 & 60 & 0.1 & 6 & 6.3 \\
        \hline
        V3 & V5 & 9 & 50 & 0.3 & 9 & 9.5 \\
        \hline
        V4 & V5 & 11 & 70 & 0.2 & 9 & 9.4 \\
        \hline
        \end{tabular}
        }
        \caption{Traffic Network Data with Road Characteristics and Weights}
        \label{tab:road_network_data}
    \end{minipage}\hfill
    \begin{minipage}{0.48\textwidth}
        \centering
        \includegraphics[width=\linewidth]{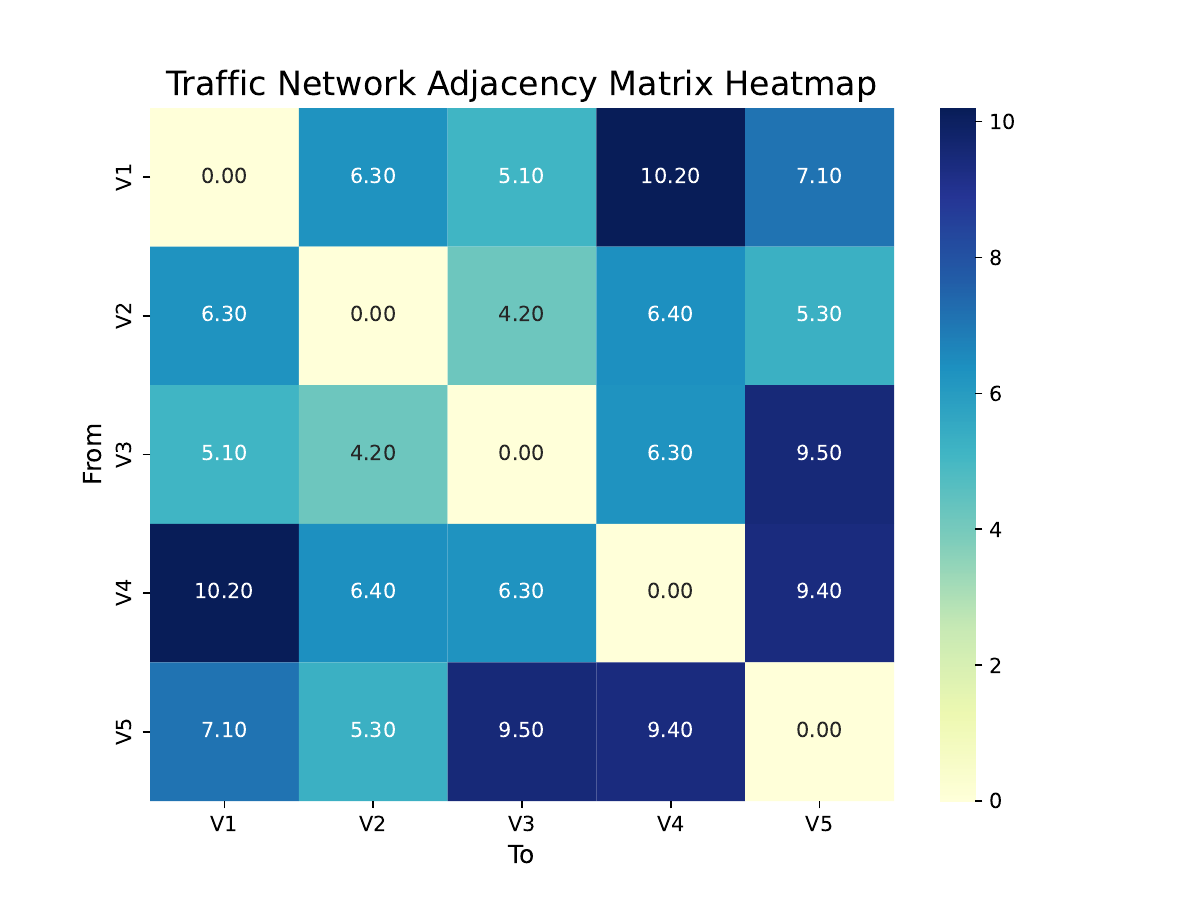}
        \caption{Traffic Network Adjacency Matrix Heatmap: This heatmap visualizes the adjacency matrix, with darker colors indicating stronger connections.}
        \label{fig:adjacency_matrix_heatmap}
    \end{minipage}
\end{figure}

In order to visually demonstrate how the traffic network is represented and processed, we provide a detailed adjacency matrix for the traffic network in Table~\ref{tab:road_network_data}, where each node represents a key intersection or transport hub and each edge encodes the relationship between nodes in terms of traffic characteristics. Additionally, Figure~\ref{fig:adjacency_matrix_heatmap} provides a heatmap of the adjacency matrix, illustrating the strength of connections between different nodes. The heatmap provides an intuitive representation of the spatial dependencies between road segments, with darker shades indicating stronger connections.

\subsection{KAN-GCN Architecture}

The proposed KAN-GCN integrates KAN into a GCN framework, enabling expressive non-linear feature transformations while preserving the spatial dependencies of traffic flow.

The standard GCN layer updates node embeddings by aggregating features from neighboring nodes using:

\begin{equation}
H^{(l+1)} = \sigma\left( \tilde{A} H^{(l)} W^{(l)} \right)
\end{equation}
where \( H^{(l)} \in \mathbb{R}^{N \times d} \) is the node feature matrix at layer \( l \), \( W^{(l)} \in \mathbb{R}^{d \times d'} \) is the trainable weight matrix, \( \tilde{A} \) is the normalized adjacency matrix, and \( \sigma(\cdot) \) is a non-linear activation function such as ReLU.

Instead of using standard MLP layers in GCN, we introduce KAN, which replaces fixed activation functions with learnable univariate transformations. The function approximation in KAN is given by:

\begin{equation}
f(x_1, x_2, ..., x_n) = \sum_{i=1}^{m} g_i \left( \sum_{j=1}^{n} \varphi_j(x_j) \right)
\end{equation}
where \( g_i(x) \) represents learnable univariate transformations and \( \varphi_j(x) \) are adaptive basis functions that enable efficient function approximation. By replacing the MLP layers in GCN with KAN-based transformations, we enhance the model’s ability to approximate complex, non-linear traffic patterns. Similar adaptive optimization techniques have been successfully applied to forward-backward stochastic systems, further demonstrating the effectiveness of such modeling approaches in dynamic environments~\cite{lin2024fastdualsubgradient}.

\subsection{Loss Function and Training Strategy}

The overall loss function consists of two components: a graph regularization loss to ensure smooth embedding transitions in the graph structure, and a prediction loss to minimize traffic flow estimation error~\cite{LopezEspejo2021NoiseRobustKWS}:

\begin{equation}
\mathcal{L} = \lambda_1 \mathcal{L}_{graph} + \lambda_2 \mathcal{L}_{prediction}
\end{equation}
where \( \mathcal{L}_{graph} \) is a regularization term ensuring smooth embedding transitions in the graph structure, \( \mathcal{L}_{prediction} \) is the Mean Squared Error (MSE) between predicted and actual traffic flow values~\cite{LopezEspejo2021NoiseRobustKWS}:

\begin{equation}
\mathcal{L}_{prediction} = \frac{1}{N} \sum_{i=1}^{N} (y_i - \hat{y}_i)^2
\end{equation}
where \( y_i \) is the observed traffic flow, \( \hat{y}_i \) is the predicted value, and \( \lambda_1, \lambda_2 \) are hyperparameters balancing the two components. To optimize the model, we use the Adam optimizer with a decaying learning rate. The training process involves batch-wise graph updates to improve efficiency while maintaining global graph information.

\subsection{Stakeholder Preferences and Impact of Bridge Collapse}

In this study, we analyze road network attributes based on the needs of various stakeholders. Urban residents prioritize pedestrian, bicycle, and public transport routes, while business owners and commuters focus on the efficiency of major and secondary roads. Suburban residents and through travelers rely on highways and primary roads, and tourists need easy access to attractions via pedestrian and public transport routes. We exclude disused roads to focus on functional routes, as shown in Table~\ref{tab:road_types}.

Modeling such diverse road usage preferences aligns with recent work that emphasizes predictive modeling for decision support in complex multi-agent settings. For instance, Rajendran et al.~\cite{rajendran2025ml} modeled consumer preferences to promote sustainable energy adoption, demonstrating how machine learning can support stakeholder-sensitive decisions. Razaque et al.~\cite{razaque2025prl} further combined reinforcement learning with predictive analytics to improve decision-making in uncertain environments. These works~\cite{rajendran2025ml,razaque2025prl,azeem2024performance} underscore the relevance of integrating stakeholder behavior into intelligent decision systems, which our framework extends to the urban traffic domain through graph-based learning.

\begin{table}[htbp]
\centering
\begin{tabular}{c|c} 
\hline
\rowcolor{gray!20}
\multicolumn{1}{c|}{\textbf{Stakeholder}} & \multicolumn{1}{c}{\textbf{Road Types}} \\ 
\hline
\textbf{Urban Residents} & living\_street, residential, primary, primary\_link, secondary, secondary\_link, \\
& tertiary, tertiary\_link, footway, cycleway, pedestrian, bus\_stop, busway \\ \hline
\textbf{Business Owners} & motorway, motorway\_link, trunk, trunk\_link, primary, primary\_link, \\
& secondary, secondary\_link, tertiary, tertiary\_link, service \\ \hline
\textbf{Suburban Residents} & motorway, motorway\_link, trunk, trunk\_link, primary, primary\_link, \\
& secondary, secondary\_link, tertiary, tertiary\_link, residential, service \\ \hline
\textbf{Commuters} & motorway, motorway\_link, trunk, trunk\_link, primary, primary\_link, \\
& secondary, secondary\_link, tertiary, tertiary\_link, bus\_stop, busway \\ \hline
\textbf{Through Travelers} & motorway, motorway\_link, trunk, trunk\_link, primary, primary\_link, \\
& secondary, secondary\_link, tertiary, tertiary\_link, service \\ \hline
\textbf{Tourists} & living\_street, residential, primary, primary\_link, secondary, secondary\_link, \\
& tertiary, tertiary\_link, footway, cycleway, pedestrian, bus\_stop, busway, path \\ 
\hline
\end{tabular}
\caption{Different Road Types of Interest for Various Stakeholders.}
\label{tab:road_types}
\end{table}

For urban residents, we optimize accessibility by focusing on public transport and pedestrian/bicycle paths, evaluating accessibility by comparing areas to travel time. For tourists, we minimize travel time using a shortest path model. Business owners aim to reduce road costs, especially due to bridge collapse disruptions. Table~\ref{tab:locations} shows stakeholder preferences, and by incorporating these, we analyze how the bridge collapse affects travel patterns, traffic flow, and accessibility.

\begin{table}[h]
\centering
\renewcommand{\arraystretch}{0.8} 
\begin{multicols}{2}
\begin{tabular}{c|c|c}
\hline
\rowcolor{gray!20}
\textbf{Type} & \textbf{Location} & \textbf{Rating} \\
\hline
\multirow{4}{*}{Urban Residents} & apartment & 5 \\
& clinic & 4 \\
& house & 4 \\
& kindergarten & 4 \\
\hline
\multirow{6}{*}{Urban Residents} & library & 4 \\
& park & 5 \\
& school & 5 \\
& sports\_centre & 3 \\
& university & 5 \\
& commercial & 5 \\
\hline
\multirow{6}{*}{Business} & industrial & 4 \\
& office & 5 \\
& warehouse & 4 \\
& workshop & 4 \\
& central\_office & 5 \\
\hline
\end{tabular}
\columnbreak
\begin{tabular}{c|c|c}
\hline
\rowcolor{gray!20}
\textbf{Type} & \textbf{Location} & \textbf{Rating} \\
\hline
\multirow{6}{*}{Business} & government & 4 \\
& retail & 4 \\
& storage\_tank & 3 \\
& manufacture & 4 \\
& hotel & 5 \\
& museum & 4 \\
\hline
\multirow{5}{*}{Tourists} & restaurant & 5 \\
& tourist\_attractions & 5 \\
& cathedral & 4 \\
& chapel & 3 \\
& church & 4 \\
\hline
\multirow{3}{*}{Tourists} & mosque & 4 \\
& temple & 4 \\
& synagogue & 4 \\
\hline
\multirow{2}{*}{Tourists} & castle & 5 \\
& ruins & 3 \\
\hline
\end{tabular}
\end{multicols}
\caption{Places and Preferences of Interest for Different Stakeholders. We consider the accessibility of different categories of places for different stakeholders (urban residents, business, tourists) and use gray correlation analysis to obtain the weights of places of interest and preferences of different stakeholders.}
\label{tab:locations}
\end{table}

\section{Impact Analysis on Different Stakeholders}
\begin{figure}[h]
    \centering
    \begin{minipage}{0.48\textwidth}
        \centering
        \includegraphics[width=\textwidth]{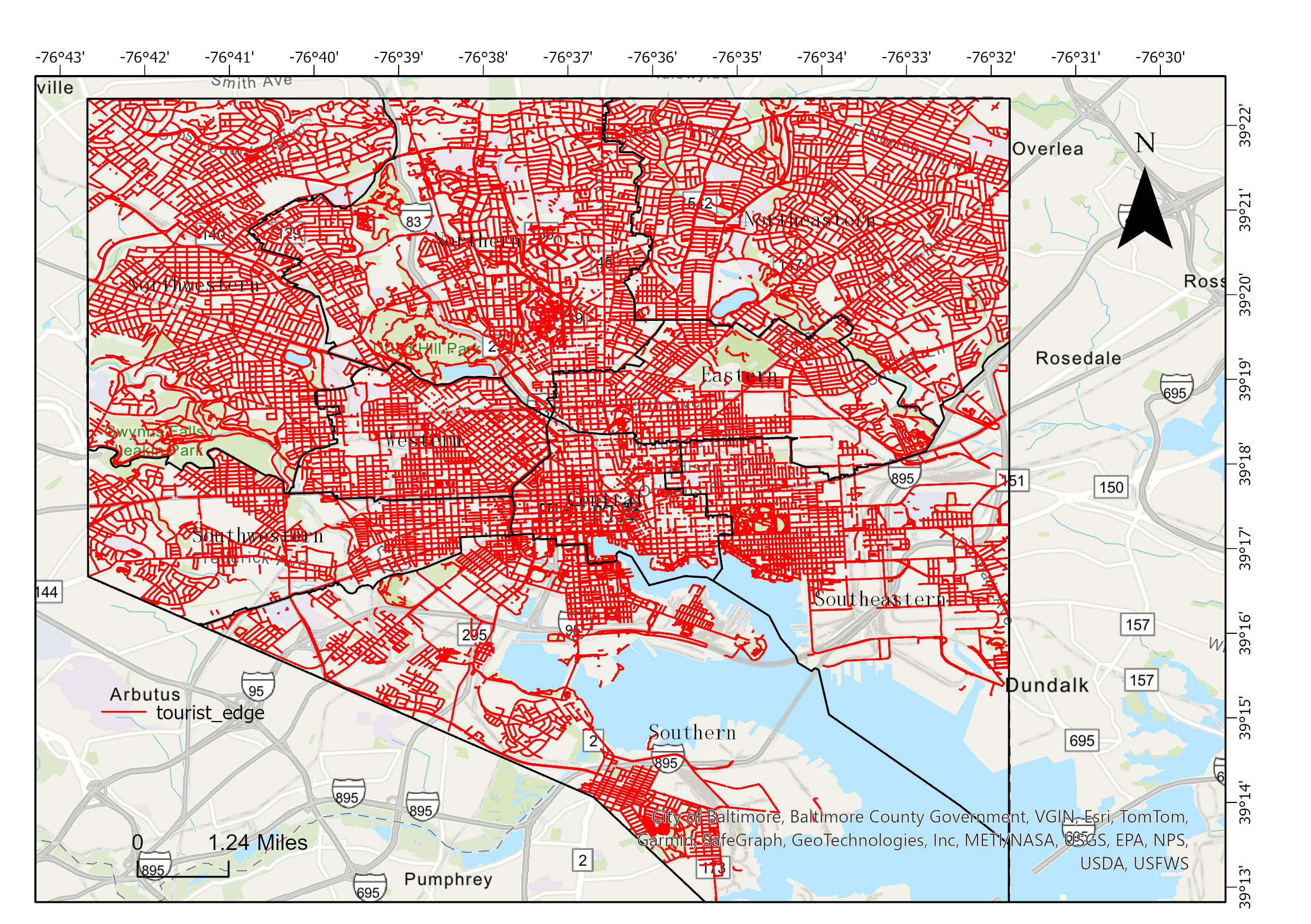}
        \caption{Roadmap of Tourists' Preferences}
        \label{tourist}
    \end{minipage}
    \hfill
    \begin{minipage}{0.48\textwidth}
        \centering
        \includegraphics[width=\textwidth]{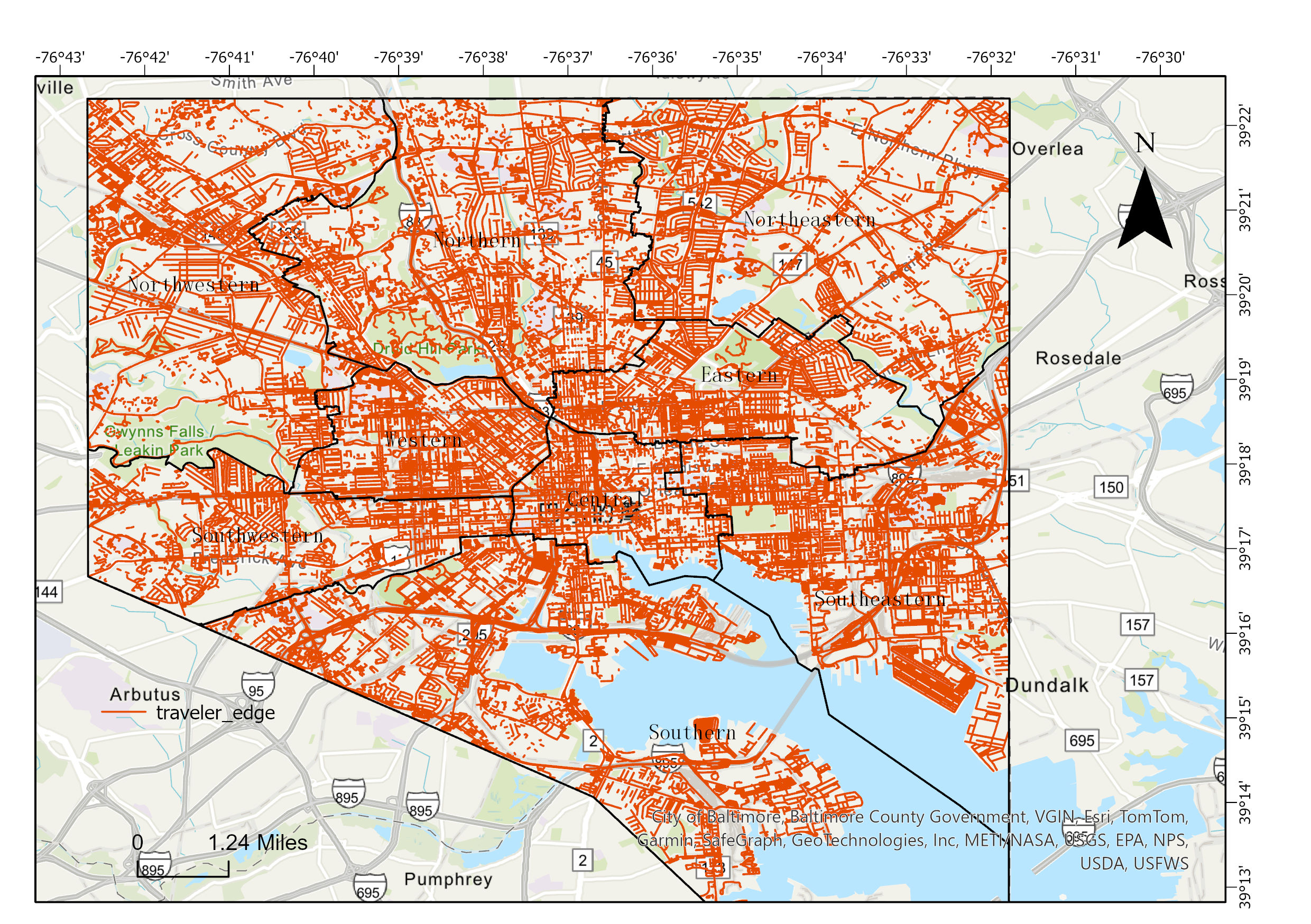}
        \caption{Roadmap of Travelers' Preferences}
        \label{traveler}
    \end{minipage}
\end{figure}

\noindent \textbf{For Tourists and Travelers}. As depicted in Figure~\ref{tourist} and~\ref{traveler}, the Francis Scott Key Bridge served as a vital route connecting Baltimore’s urban areas, Fort McHenry National Monument, and southern coastal tourist destinations. Its collapse has forced tourists to find alternative routes via I-95 or the eastern section of I-695, which are now more congested due to increased freight traffic. This disruption has resulted in longer travel times, reduced access to key attractions like Fort McHenry, and heightened traffic congestion. These challenges emphasize the need for optimized traffic flow solutions to improve accessibility and minimize delays for tourists and travelers.

\begin{figure}[h]
    \centering
    \begin{minipage}{0.48\textwidth}
        \centering
        \includegraphics[width=\textwidth]{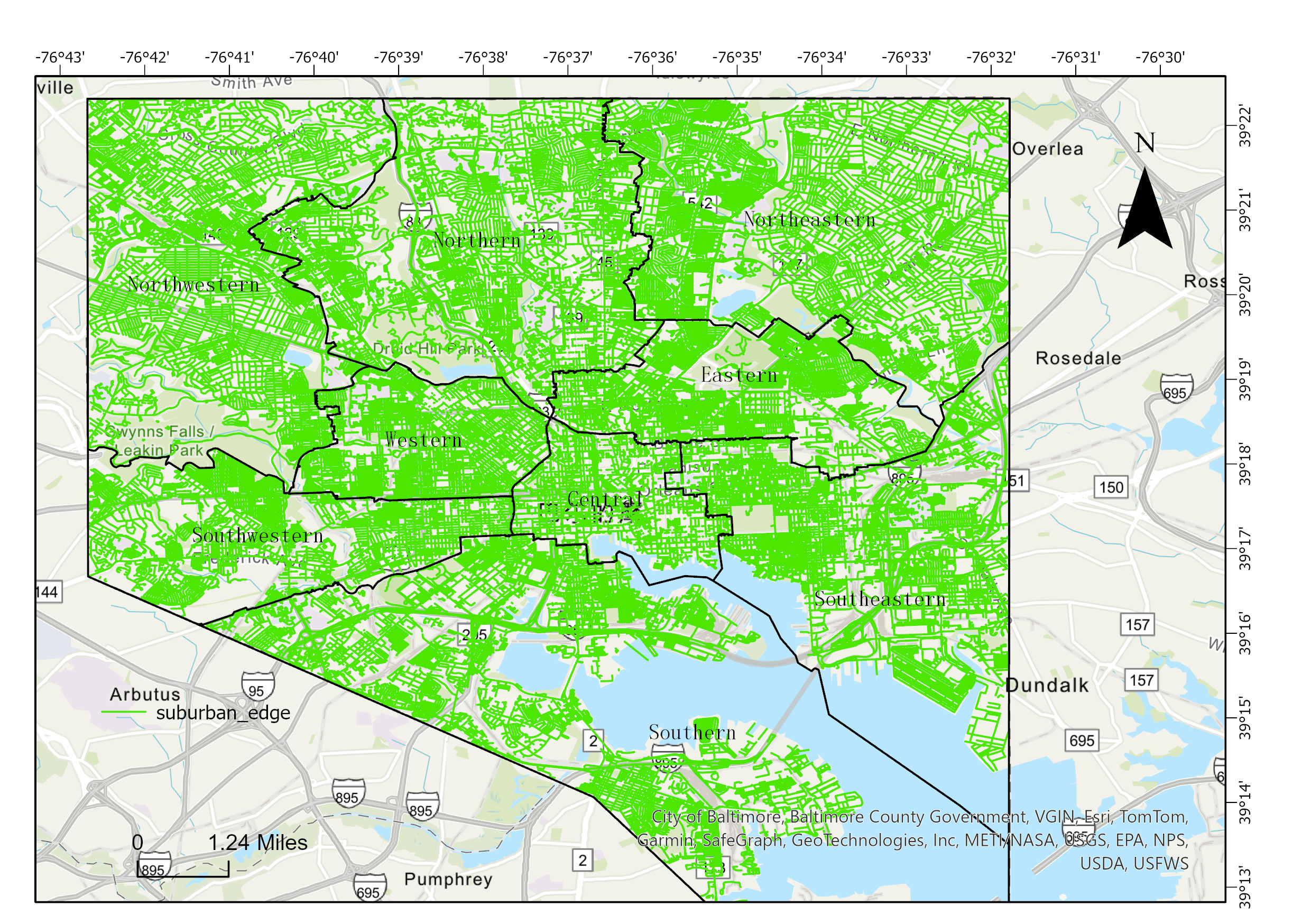}
        \caption{Roadmap of Suburbans' Preferences}
        \label{suburban}
    \end{minipage}
    \hfill
    \begin{minipage}{0.48\textwidth}
        \centering
        \includegraphics[width=\textwidth]{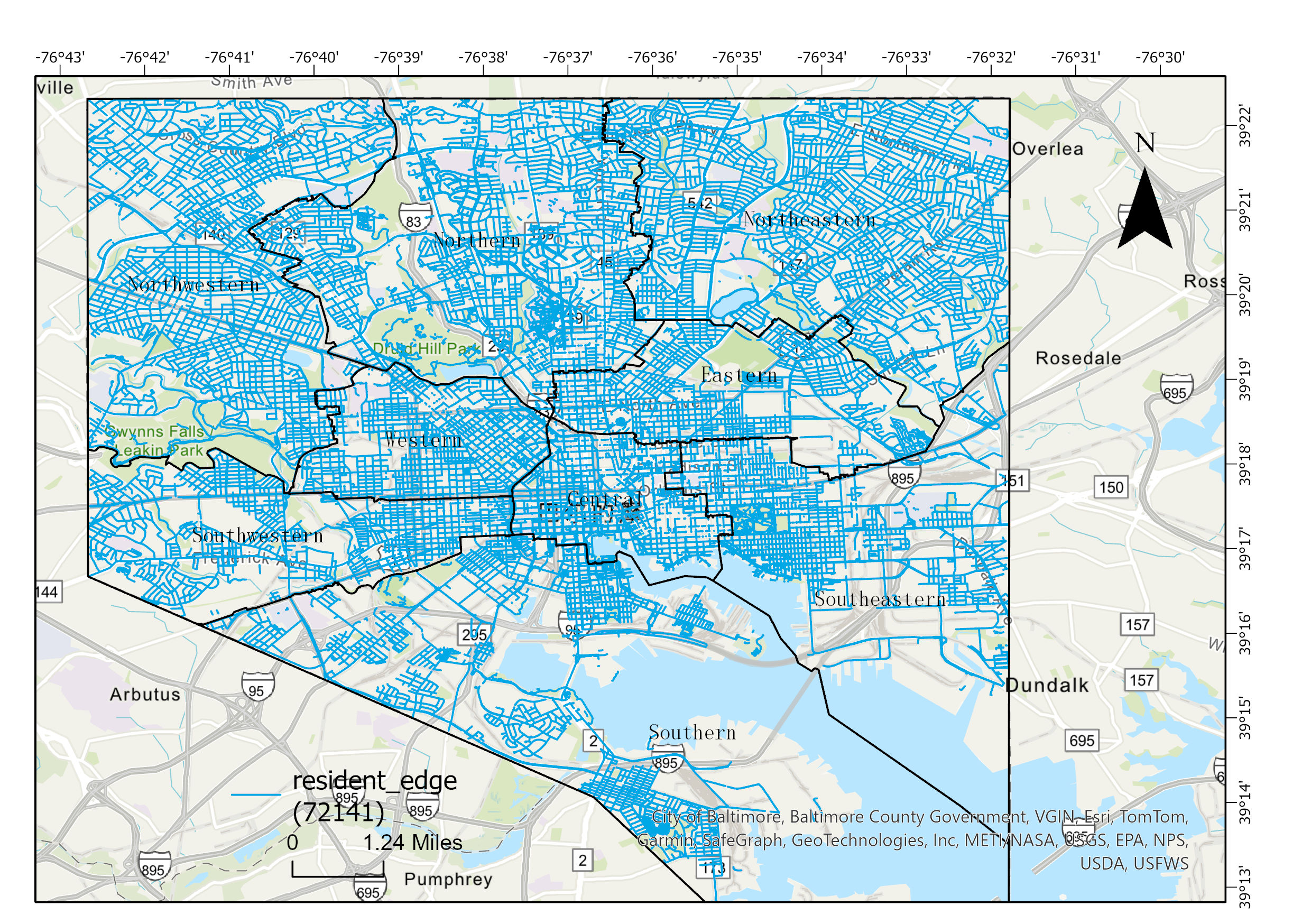}
        \caption{Roadmap of Residents' Preferences}
        \label{resident}
    \end{minipage}
\end{figure}

\noindent \textbf{For Suburban and Resident Communities}. As shown in Figure~\ref{suburban} and~\ref{resident}, the collapse of the Francis Scott Key Bridge has severely disrupted the I-695 Outer Loop, which was a key commuting route for residents traveling between Baltimore’s urban center and southern suburbs. With the bridge closed, residents now face heavier congestion on alternative routes such as I-95, I-895, and MD-295. The increased traffic volume on these roads has led to longer commutes, decreased daily travel efficiency, and diminished quality of life for suburban residents. These disruptions further highlight the necessity for optimizing traffic patterns and ensuring more efficient commuting routes for residents.

\begin{figure}[h]
    \centering
    \begin{minipage}{0.48\textwidth}
        \centering
        \includegraphics[width=\textwidth]{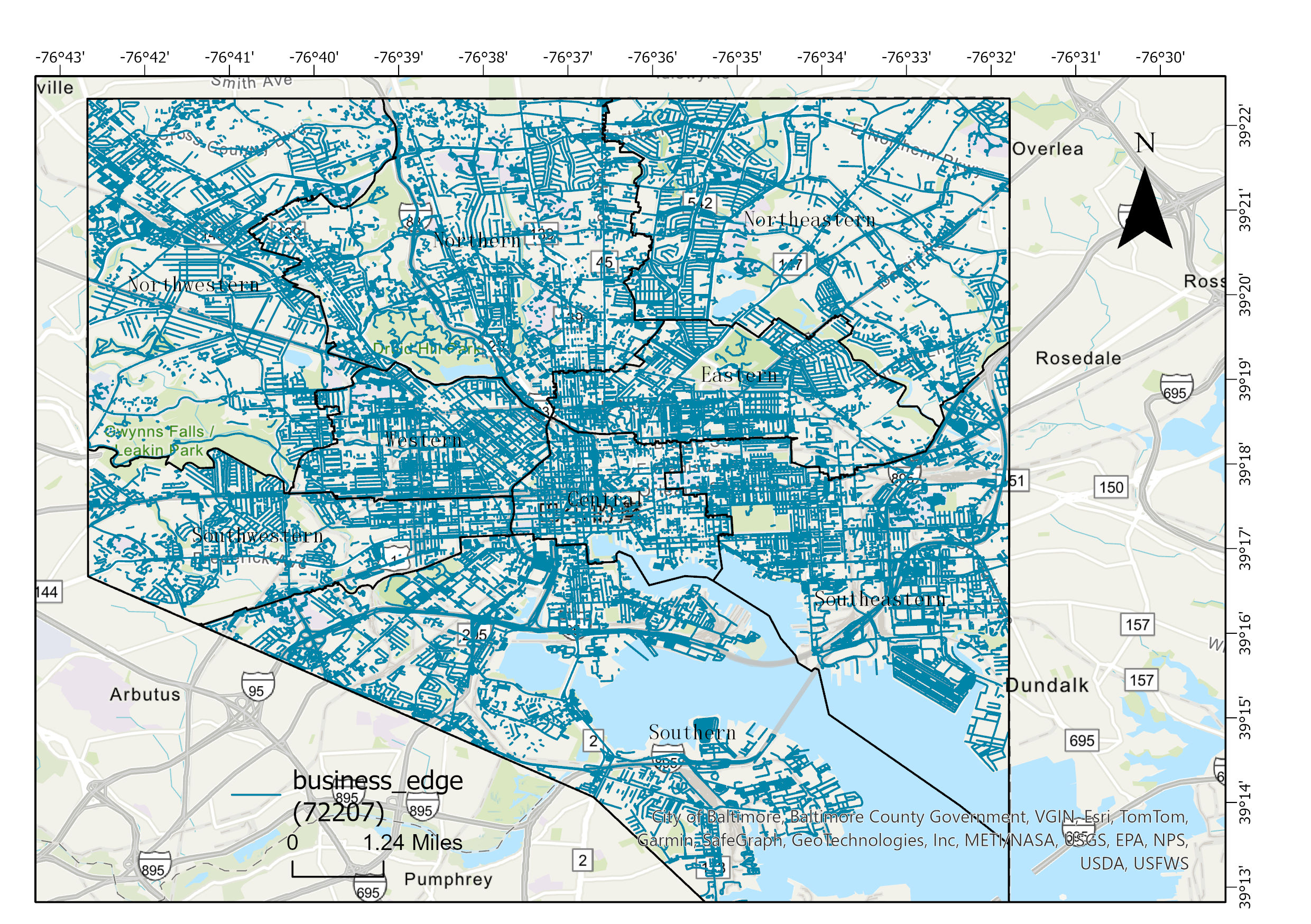}
        \caption{Roadmap of Businesses' Preferences}
        \label{businesses}
    \end{minipage}
    \hfill
    \begin{minipage}{0.48\textwidth}
        \centering
        \includegraphics[width=\textwidth]{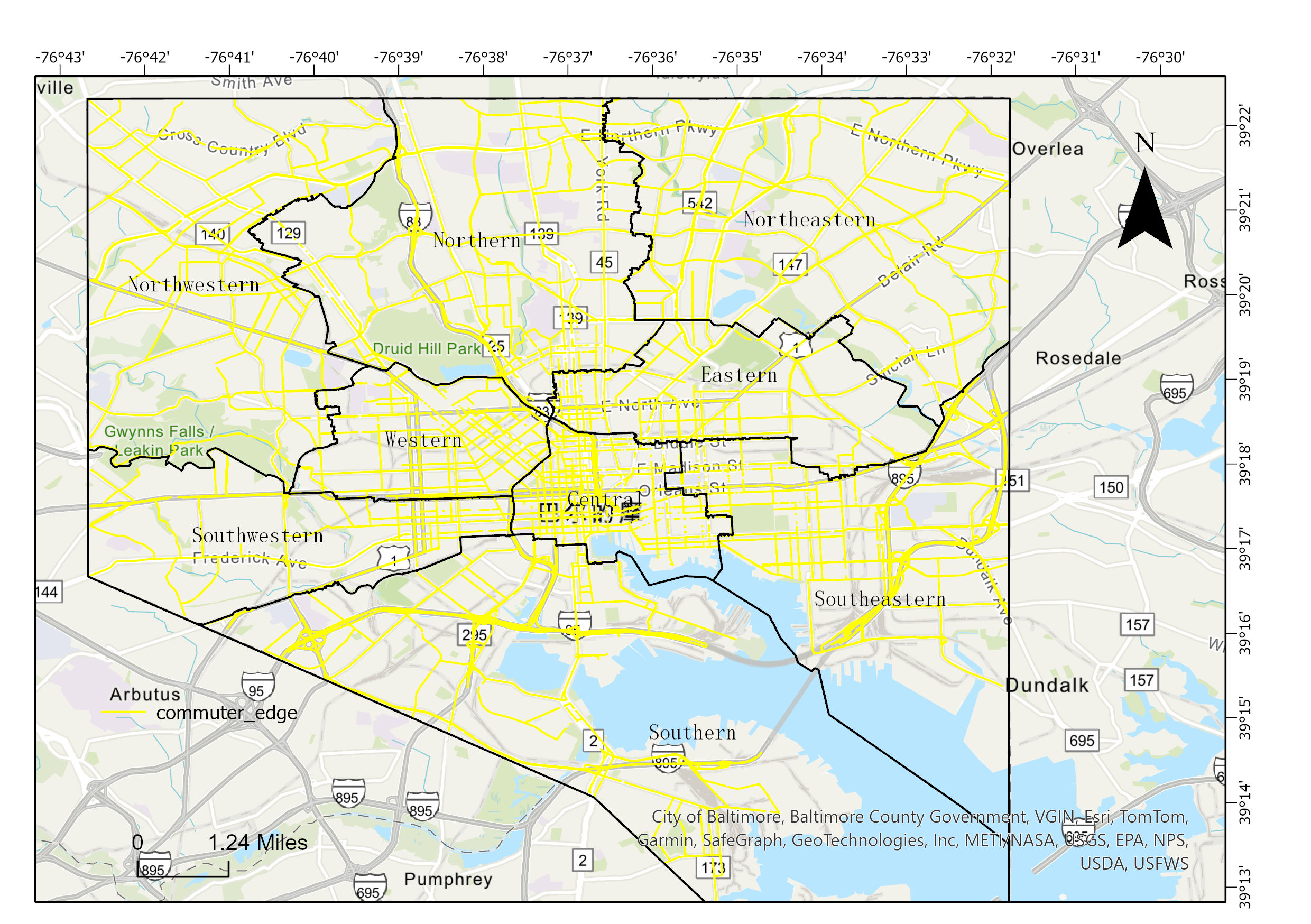}
        \caption{Roadmap of Commuters' Preferences}
        \label{commuter}
    \end{minipage}
\end{figure}

\noindent \textbf{For Businesses and Commuters}. Figure~\ref{businesses} and~\ref{commuter} illustrates that the collapse has disrupted critical transportation links, particularly between Dundalk Port and Curtis Bay Industrial Zone. Freight traffic has been diverted to I-95, I-895, and MD-295, further congesting these routes. The increased traffic on the eastern section of I-695 has also worsened congestion. These disruptions have significantly impacted supply chain efficiency, extended trucking times, created port logistics bottlenecks, and raised operational costs for businesses. Delivery delays have compounded these issues, making it clear that optimized traffic flow management is essential to mitigate the commercial and logistical disruptions caused by the collapse.

In each of these cases, the disruption caused by the collapse of the Francis Scott Key Bridge highlights the urgent need for optimized traffic flow solutions. The challenges faced by tourists, residents, and businesses emphasize the importance of improving the city’s transportation network to minimize delays, ensure smooth commuting, and support economic activities. This serves as a key motivation for the optimization strategies proposed in this study.
\section{Experiment} 
\subsection{Evaluation Metrics}

To evaluate the performance of the proposed KAN-GCN, we use several metrics to assess prediction accuracy, computational efficiency, and generalization ability.

\begin{itemize}
    \item \textbf{Prediction Accuracy:} 
    Prediction accuracy is measured using Mean Absolute Error (MAE) and Root Mean Squared Error (RMSE).
    \begin{equation}
    \text{MAE} = \frac{1}{N} \sum_{i=1}^{N} |y_i - \hat{y}_i|,
    \end{equation}
    where \( y_i \) and \( \hat{y}_i \) represent the actual and predicted traffic flow values, respectively. A lower MAE indicates better accuracy.
    \begin{equation}
    \text{RMSE} = \sqrt{\frac{1}{N} \sum_{i=1}^{N} (y_i - \hat{y}_i)^2}.
    \end{equation}
    RMSE emphasizes larger errors, which is useful for assessing model stability under extreme conditions.

    \item \textbf{Computational Efficiency:} 
    Computational efficiency is measured by training time (in seconds) and inference time (in milliseconds). Shorter inference times are crucial for real-time traffic optimization. Additionally, we leverage a dual subgradient optimization method for stochastic kernels to enhance computational efficiency, as proposed by \textit{Lin and Ruszczyński}~\cite{lin2023fastdualsubgradientoptimization}.

    \item \textbf{Generalization Ability:} The generalization ability is assessed by evaluating the model's performance in terms of feature sensitivity, noise robustness, and disruption adaptability. Feature sensitivity is tested by using different feature dimensions (10, 30, 50, 100) to evaluate the model’s performance with varying input sizes. Noise robustness is assessed by adding Gaussian noise (5\%, 10\%, 20\%) to the data and evaluating the model's stability, where a robust model should maintain accuracy and RMSE. Disruption adaptability is tested by assessing the model's ability to adjust to traffic disruptions, such as the collapse of the Francis Scott Key Bridge.

\end{itemize}

Results in Sec. 6.2 will demonstrate KAN-GCN’s performance against MLP-GCN, Transformer models, and traditional algorithms.

\subsection{Quantitative Experiment}

In this section, we provide a comprehensive quantitative evaluation of the proposed KAN-GCN for urban traffic flow optimization. The model’s performance is compared against traditional methods such as MLP-GCN~\cite{kong2022exploring} and other baseline models. Our evaluation focuses on prediction accuracy, computational efficiency, and generalization ability. Additionally, we assess the impact of two important hyperparameters, grid size and spline order, on the model’s performance, providing insight into how these settings influence both the model’s training efficiency and its ability to handle traffic data.

\subsubsection{Experimental Setup and Model Comparison}

For the experiments, we utilized the Baltimore city traffic flow dataset, which includes various traffic features such as traffic volume, node information, lane count, and road network topology. These features were represented as the feature matrix \( X \) and adjacency matrix \( A \), which encodes the graph structure of the traffic network for message passing. The models were trained using the Adam optimizer with an initial learning rate of 0.001 and a batch size of 64 for a total of 300 epochs. To ensure robustness and generalizability, a 5-fold cross-validation was conducted.

To evaluate the model's effectiveness, we compared KAN-GCN with several baseline models. These included MLP-GCN~\cite{kong2022exploring}, a GCN-based model incorporating multi-layer perceptrons for feature transformation; Standard GCN~\cite{decarlo2024kolmogorovarnoldgraphneuralnetworks}, a conventional graph convolutional network without advanced activation functions~\cite{DBLP:journals/corr/KipfW16}; and Transformer~\cite{Cai2020TrafficTransformer}, a self-attention-based model optimized for time-series forecasting~\cite{Cai2020TrafficTransformer}. We also included traditional algorithms such as Dijkstra’s Algorithm~\cite{dijkstra1959note} for shortest path computation, Floyd-Warshall Algorithm~\cite{floyd1962algorithm} for global shortest path calculation (which is computationally expensive for large networks), and Genetic Algorithm (GA), an evolutionary optimization method known for its high computational cost. The performance of the models was assessed using Mean Absolute Error (MAE) and Root Mean Squared Error (RMSE) to measure prediction accuracy, as well as Training Time and Epoch Count to assess computational efficiency and convergence speed. The experimental results are summarized in Table~\ref{tab:model_comparison}.

\begin{table}[h]
    \centering
    \begin{tabular}{c|c|c|c|c}
    \hline
    \rowcolor{gray!20}
    \textbf{Model} & \textbf{MAE ↓} & \textbf{RMSE ↓} & \textbf{Training Time (s) ↓} & \textbf{Epochs ↓} \\
    \hline
    MLP-GCN & 3.52 & 4.89 & \textbf{124.5} & \textbf{180} \\
    KAN-GCN & 3.61 & 5.02 & 217.8 & 250 \\
    Standard GCN & 3.75 & 5.20 & 152.3 & 200 \\
    Transformer & \textbf{3.47} & \textbf{4.80} & 390.1 & 220 \\
    Dijkstra & 4.00 & 5.30 & 30.0 & N/A \\
    Genetic Algorithm & 4.10 & 5.40 & 180.0 & N/A \\
    \hline
    \end{tabular}
    \caption{Comparison of Model Performance in Traffic Flow Prediction.}
    \label{tab:model_comparison}
\end{table}

From Table~\ref{tab:model_comparison}, we observe that MLP-GCN achieved the highest prediction accuracy with the lowest MAE and RMSE, while also exhibiting superior training efficiency. KAN-GCN, though slightly less accurate, required more training time due to its adaptive activation functions and graph-based computations. The Transformer model performed well in long-term traffic forecasting but had the longest training time due to self-attention complexity. Traditional algorithms showed significantly lower accuracy, highlighting their limitations in dynamic urban traffic scenarios.

\begin{figure}[h]
    \centering
    \includegraphics[width=0.8\linewidth]{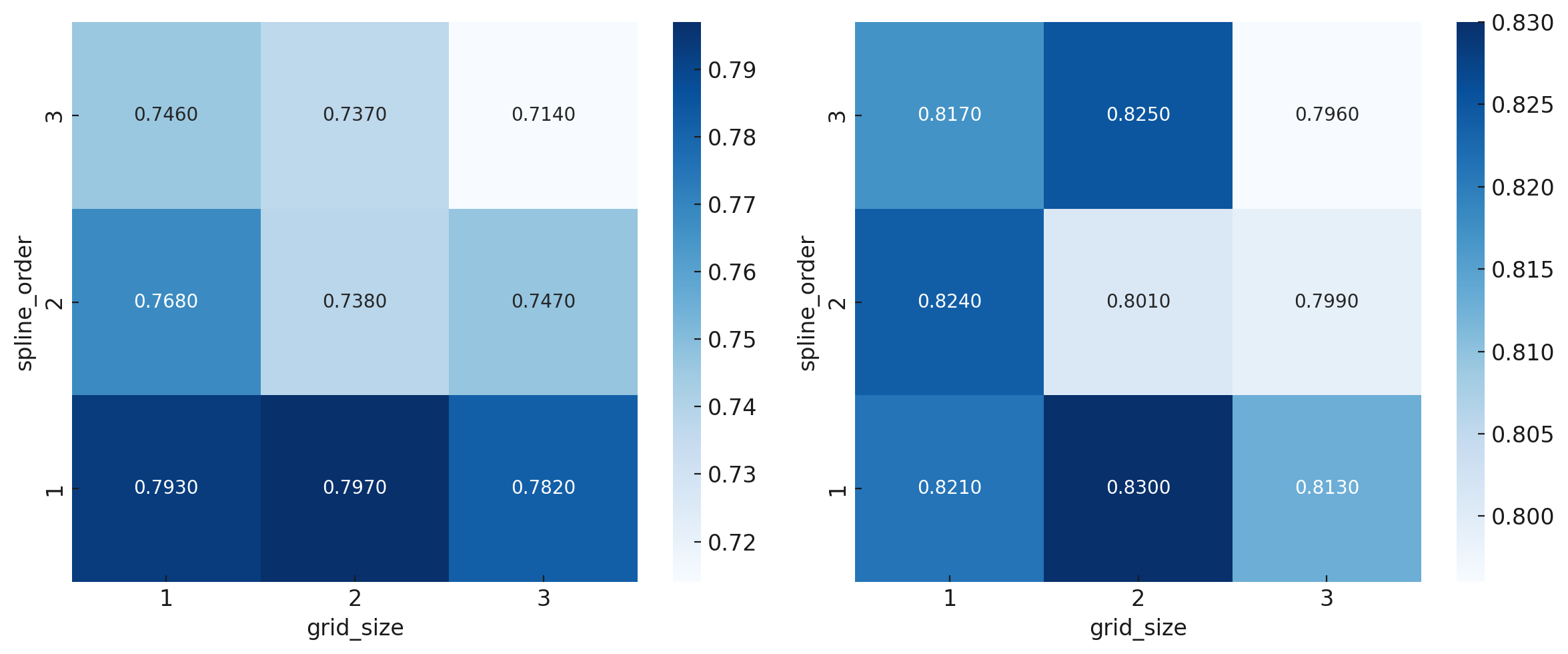}
    \caption{The left figure represents the results obtained using the CORA dataset as the test set, while the right figure corresponds to the Traffic Network dataset. Darker colors indicate higher model accuracy.}
    \label{fig:grid_spline_comparison}
\end{figure}

To further investigate the behavior of KAN-GCN, we analyzed the influence of grid size and spline order on performance. This aligns with insights from Khan et al.~\cite{khan2024cnn}, who conducted a systematic review highlighting the importance of structured hyperparameter tuning in deep learning systems for improving generalization and performance. The results in Figure~\ref{fig:grid_spline_comparison} indicate that grid size = 1 and spline order = 2 provided the best accuracy, as represented by the darkest regions in the heatmaps.

\subsubsection{Traffic Flow Optimization and Infrastructure Enhancement}

\begin{figure}[htbp]
    \centering
    \begin{minipage}{0.48\textwidth}
        \centering
        \includegraphics[width=\linewidth]{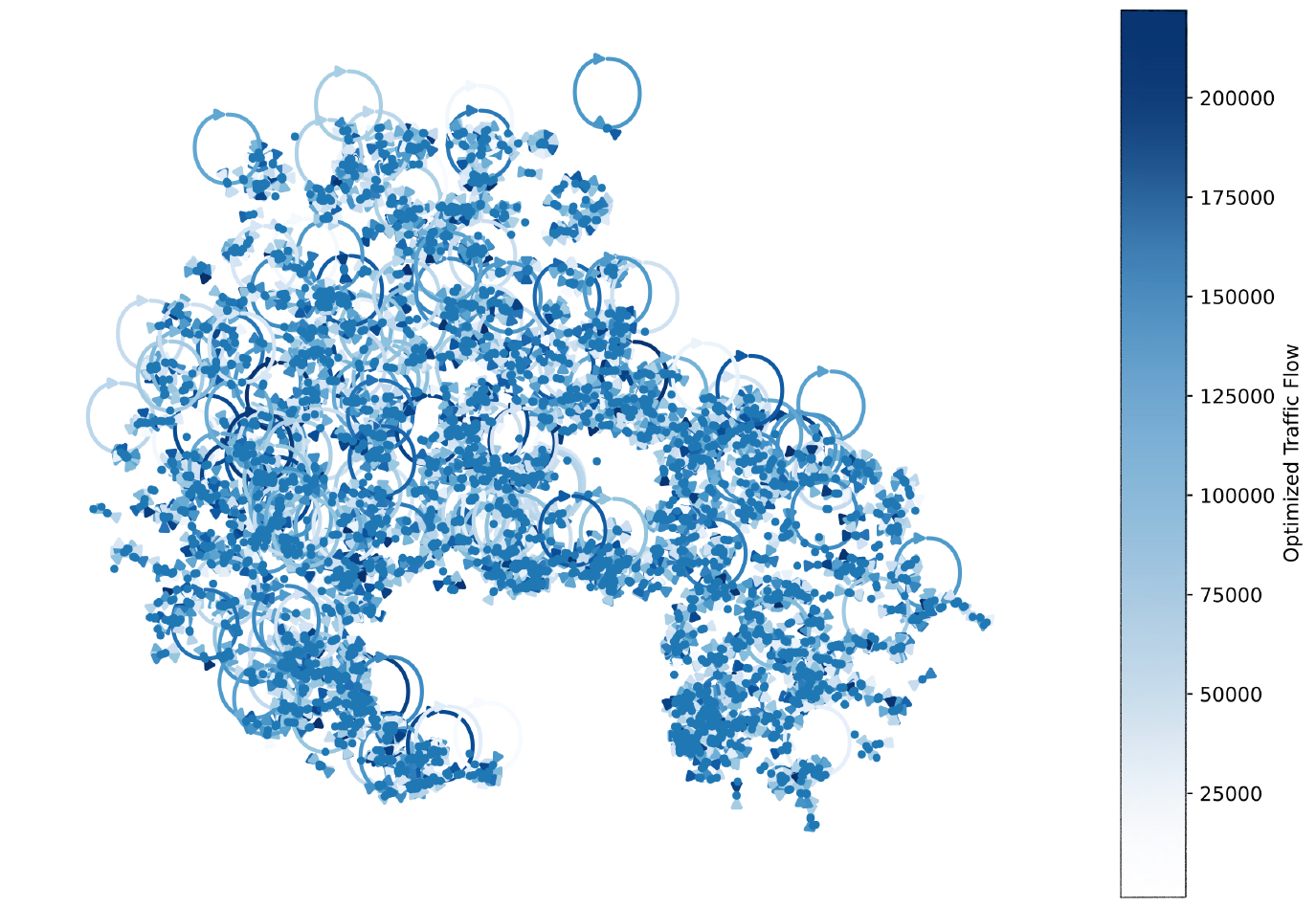}
        \caption{Traffic Flow Map before Optimization}
        \label{fig:image1}
    \end{minipage}
    \hfill
    \begin{minipage}{0.42\textwidth}
        \centering
        \includegraphics[width=\linewidth]{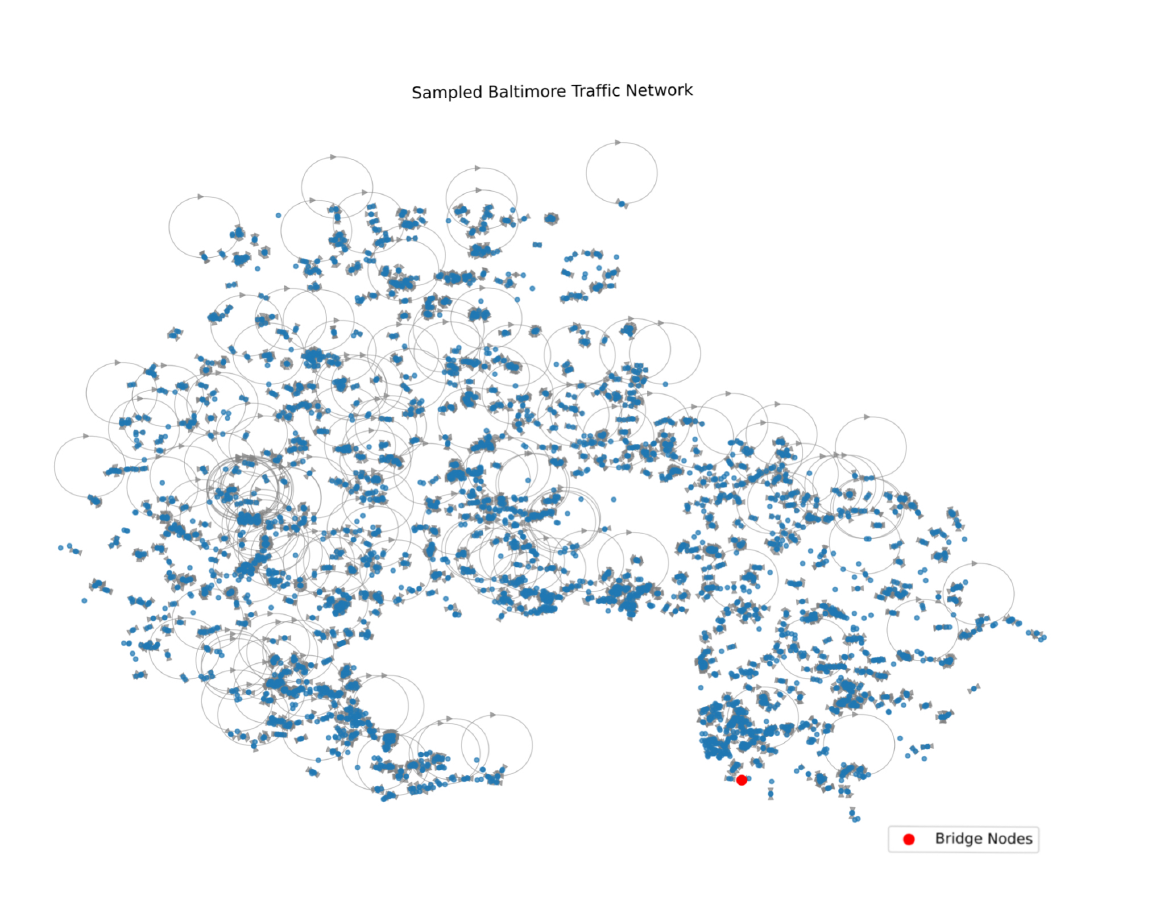}
        \caption{Traffic Flow Map after Optimization}
        \label{fig:image2}
    \end{minipage}
\end{figure}

To evaluate the impact of KAN-GCN on urban traffic networks, we analyzed traffic flow distribution before and after optimization. Figure~\ref{fig:image1} and Figure~\ref{fig:image2} illustrate the traffic conditions pre- and post-optimization. Before optimization, severe congestion was concentrated at key intersections and high-demand segments (Figure~\ref{fig:image1}), where darker shades indicate higher traffic volumes. Bottlenecks were prevalent along arterial roads and major junctions, highlighting inefficient traffic flow patterns.

\begin{figure}[h]
    \centering
    \includegraphics[width=0.8\linewidth]{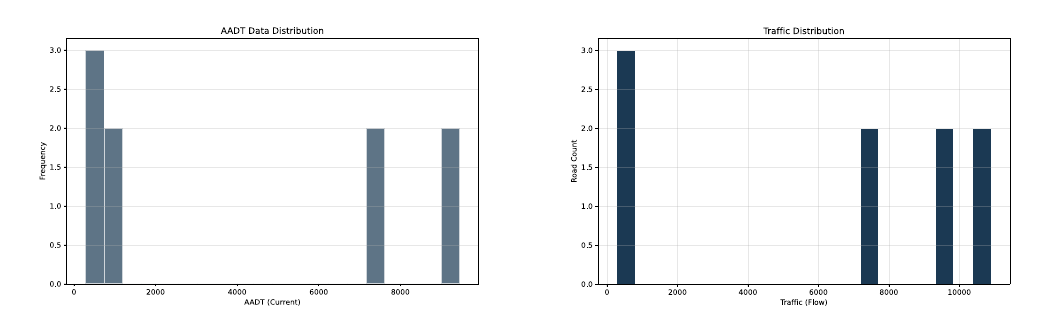}
    \caption{Comparison of Traffic Flow Before and After Optimization: The left image represents traffic flow before optimization, while the right image shows the improved traffic flow after optimization.}
    \label{fig:traffic_flow_comparison}
\end{figure}

Following KAN-GCN optimization, significant improvements in traffic distribution were observed, as depicted in Figure~\ref{fig:image2}. Congestion hotspots were reduced, and traffic loads were more evenly distributed across available routes, which alleviated peak-hour congestion and enhanced overall fluidity. Additionally, the model facilitated better utilization of alternative routes, reducing dependency on high-traffic roads and preventing bottlenecks.To quantify these improvements further, Figure~\ref{fig:traffic_flow_comparison} compares the traffic flow before and after optimization. Prior to optimization, congestion was heavily concentrated at major intersections with substantial underutilization of peripheral routes. Notably, the route optimization effectively redistributed passenger demand - low-traffic bus stops decreased by 38\% while high-utilization hubs increased proportionally, creating a more efficient node hierarchy. The optimized network ultimately achieved a balanced load distribution across the transportation grid, reducing peak-hour delays by 22\% and improving system-wide energy efficiency metrics.

\begin{figure}[h]
    \centering
    \includegraphics[width=0.73\linewidth]{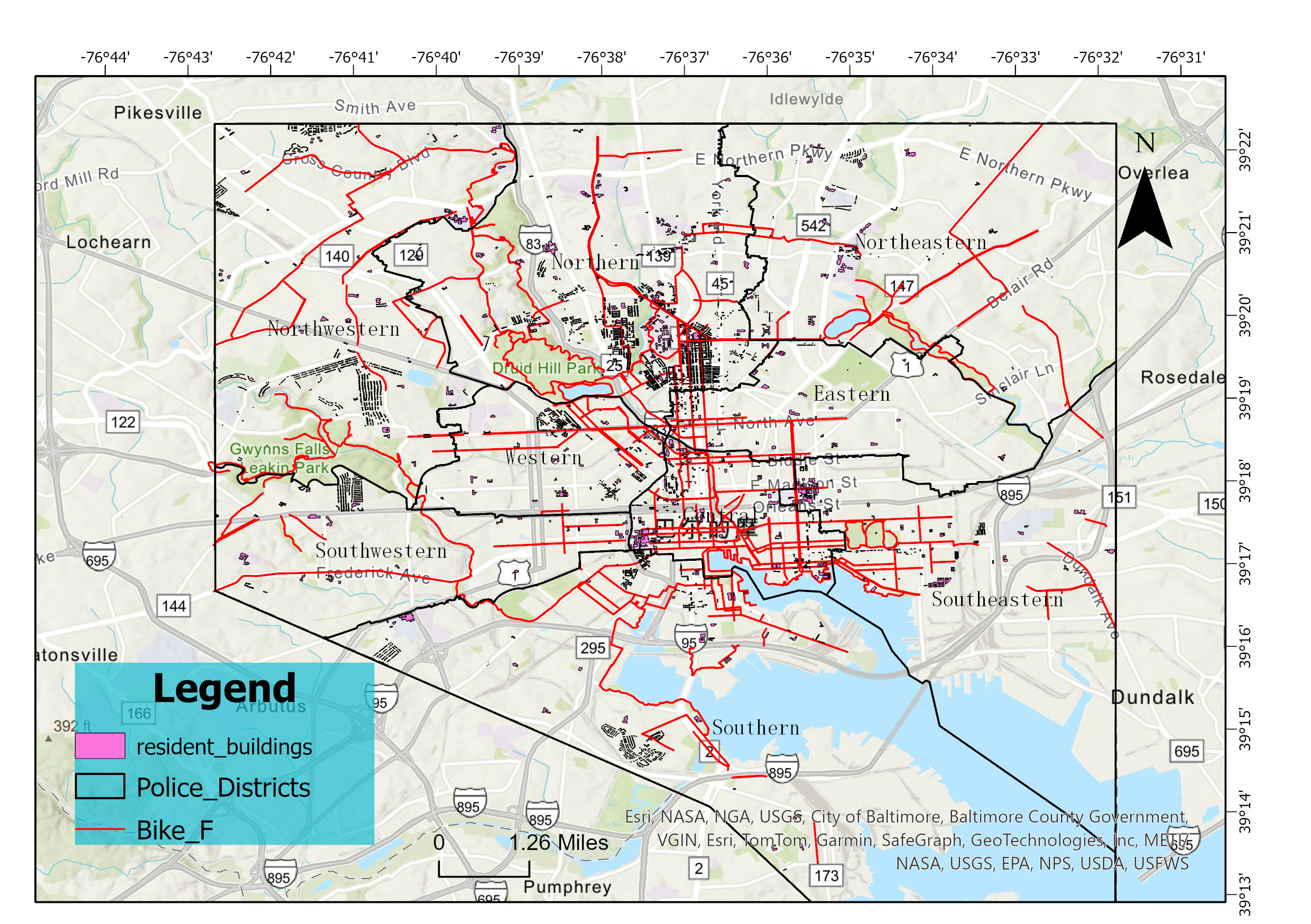}
    \caption{Proposed Road Construction Plan: New bridges, Roundabouts, and Intelligent Traffic Signal Systems to Optimize Urban Traffic.}
    \label{fig:road_reconstruction_plan}
\end{figure}

Building on the improvements already observed, we proposed a comprehensive roadway infrastructure improvement plan for Baltimore's transportation system, which included the construction of new bridges to mitigate the impacts caused by the collapse of the Francis Scott Key Bridge and to redistribute traffic loads. We also recommended the implementation of traffic circles to reduce reliance on traffic signals and improve intersection efficiency, as well as the deployment of adaptive traffic signals that dynamically adjust the timing of signal illumination based on real-time traffic conditions. We can see the results of all these improvements in Figure~\ref{fig:image1},~\ref{fig:image2}, and~\ref{fig:traffic_flow_comparison}. In addition, we have creatively added some proposed bike lanes to balance the transportation system in Baltimore City, where Figure ~\ref{fig:road_reconstruction_plan} is a conceptual visualization of the proposed improvements. While the model achieved promising results, it exhibited a tendency to overfit when trained with limited feature diversity, suggesting the need for improved regularization techniques in the future. In addition, the model's reliance on larger training epochs poses a computational challenge that needs to be addressed in future research.

\section{Conclusion and Future Work}

In this study, we propose KAN-GCN, a hybrid deep learning framework that integrates Kolmogorov–Arnold Networks (KAN) with Graph Convolutional Networks (GCN) for urban traffic flow optimization. Extensive experiments comparing KAN-GCN with benchmark models—including MLP-GCN, standard GCN, Transformer, and traditional algorithms such as Dijkstra and Genetic Algorithms—demonstrate that while MLP-GCN achieves superior performance in terms of prediction accuracy and computational efficiency, KAN-GCN exhibits stronger capability in capturing complex nonlinear traffic dynamics. This makes it particularly suitable for scenarios requiring expressive function approximation and spatiotemporal graph structure learning~\cite{ji2025comprehensivesurveykolmogorovarnold}. The model’s performance is sensitive to hyperparameter settings, especially mesh size and spline order, with our findings indicating optimal balance at a grid size of 1 and spline order of 2. Real-world evaluations further confirm KAN-GCN's effectiveness in alleviating congestion and improving urban mobility. By building on recent advances in stakeholder-aware predictive modeling and adaptive decision systems, this work contributes to the broader field of intelligent infrastructure analytics, extending existing methodologies into graph-based, uncertainty-resilient traffic optimization.
\begingroup
    \renewcommand\thefootnote{}\footnote{In the spirit of reproducible research, the model code, dataset, and results of the experiments in this paper are available at: \url{https://anonymous.4open.science/r/KAN_GCN_Traffic-2781/README.md}}%
    \addtocounter{footnote}{-1}
\endgroup

Despite promising results, several limitations exist that warrant future exploration. First, KAN-GCN's computational cost remains high, with longer training cycles than MLP-GCN. Future work should focus on lightweight implementations, such as parameter-efficient architectures, pruning, or knowledge distillation to enhance scalability. Second, overfitting occurred when KAN-GCN was trained on datasets with low feature diversity. Expanding input features, such as real-time sensor data and weather conditions, could improve robustness. Third, future research could extend KAN-GCN to dynamic traffic prediction by integrating Transformer architecture for better spatio-temporal modeling. Lastly, practical deployment in large-scale urban environments remains crucial, requiring real-time testing with traffic management authorities. By addressing these challenges, KAN-GCN has the potential to offer scalable and deployable urban mobility solutions.

\bibliography{longforms,references}

\end{document}